\documentclass[10pt,twocolumn,letterpaper]{article}

\usepackage{wacv}              

\usepackage{graphicx}
\usepackage{amsmath}
\usepackage{amssymb}
\usepackage{booktabs}
\usepackage{array}
\usepackage{booktabs}
\usepackage{multirow}
\usepackage[normalem]{ulem}
\useunder{\uline}{\ul}{}
\usepackage[vlined,ruled]{algorithm2e}

\usepackage{cuted}
\usepackage{appendix}

\newcommand{\T}{\mathsf{\!T}}

\usepackage[pagebackref,breaklinks,colorlinks]{hyperref}

\usepackage[capitalize]{cleveref}
\crefname{section}{Sec.}{Secs.}
\Crefname{section}{Section}{Sections}
\Crefname{table}{Table}{Tables}
\crefname{table}{Tab.}{Tabs.}

\def\wacvPaperID{1402} 
\def\confName{WACV}
\def\confYear{2025}

\begin{document}

\title{VADet: Multi-frame LiDAR 3D Object Detection~using Variable Aggregation}

\author{Chengjie Huang \qquad Vahdat Abdelzad \qquad Sean Sedwards \qquad Krzysztof Czarnecki \\
University of Waterloo\\
{\tt\small \{c.huang,vahdat.abdelzad,sean.sedwards,k2czarne\}@uwaterloo.ca}}
\maketitle

\begin{abstract}
Input aggregation is a simple technique used by state-of-the-art LiDAR 3D object detectors to improve detection. However, increasing aggregation is known to have diminishing returns and even performance degradation, due to objects responding differently to the number of aggregated frames. To address this limitation, we propose an efficient adaptive method, which we call Variable Aggregation Detection (VADet). Instead of aggregating the entire scene using a fixed number of frames, VADet performs aggregation per object, with the number of frames determined by an object's observed properties, such as speed and point density.
VADet thus reduces the inherent trade-offs of fixed aggregation and is not architecture specific. To demonstrate its benefits, we apply VADet to three popular single-stage  detectors and achieve state-of-the-art performance on the Waymo dataset.
\end{abstract}


\section{Introduction}
\label{sec:intro}

\begin{figure}[t]
    \centering
    \begin{subfigure}[b]{\linewidth}
         \centering
         \includegraphics[width=\textwidth]{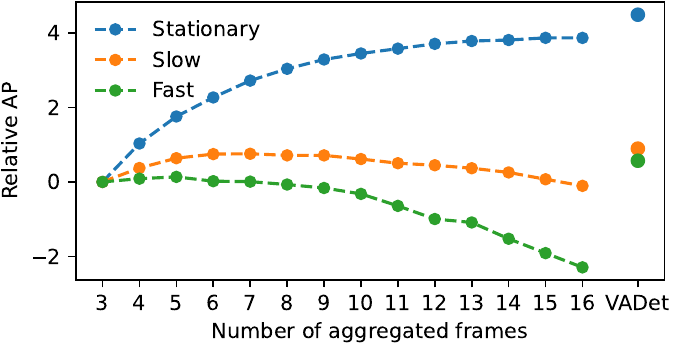}
         \caption{Vehicle AP performance}
         \label{fig:static-dynamic-aph}
     \end{subfigure}
     \begin{subfigure}[b]{0.3\linewidth}
         \centering
         \includegraphics[width=\textwidth]{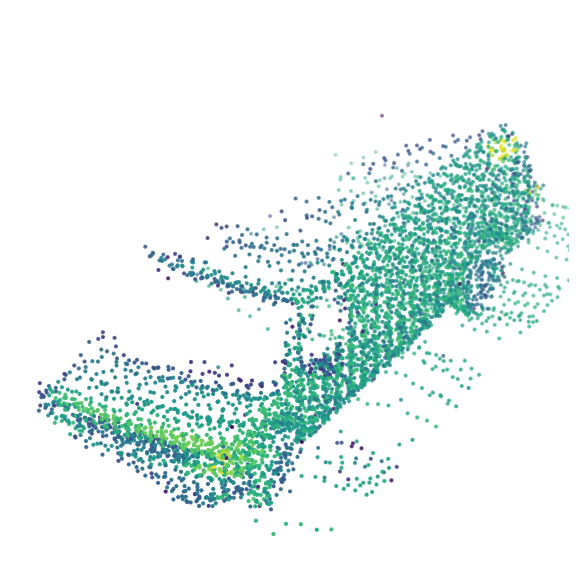}
         \vspace{0.1cm}
         \caption{Stationary vehicle}
         \label{fig:static-dynamic-static}
     \end{subfigure}
     \hspace{2em}
     \begin{subfigure}[b]{0.3\linewidth}
         \centering
         \includegraphics[width=\textwidth]{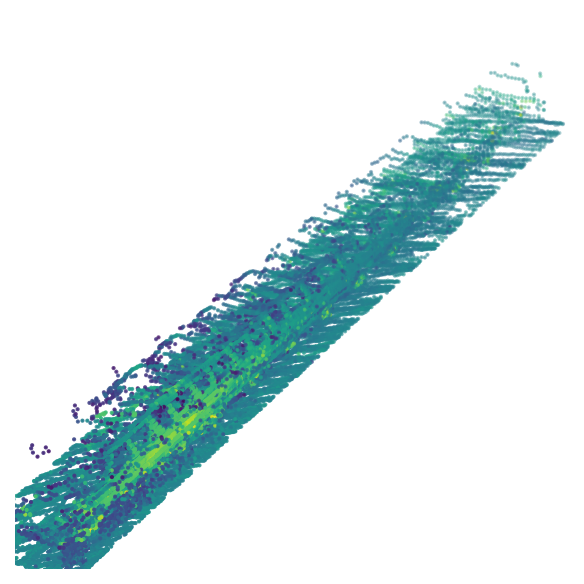}
         \vspace{0.1cm}
         \caption{Fast vehicle}
         \label{fig:static-dynamic-dynamic}
     \end{subfigure}
    \caption{Performance trade-off between stationary ($<$0.2\,m/s), slow ([0.2,10)\,m/s), and fast (10\,m/s) vehicles. The relative AP in~\subref{fig:static-dynamic-aph} illustrates the improvement/degradation relative to using 3-frame fixed aggregation. \subref{fig:static-dynamic-static} and \subref{fig:static-dynamic-dynamic} are examples of stationary and fast-moving vehicles after 16-frame fixed aggregation.}
    \label{fig:static-dynamic}
    \vspace{-1em}
\end{figure}

LiDAR-based methods give state-of-the-art (SOTA) 3D object detection performance in autonomous driving. While object detectors can produce accurate detections from a single LiDAR point cloud~\cite{chen2023voxelnext,wang2023dsvt}, they have been shown to benefit from aggregated input consisting of multiple consecutive frames. A widely adopted method for incorporating multi-frame sequential point clouds is what we term fixed aggregation, where a fixed number of frames are concatenated after ego motion correction, often including timestamps as an additional feature. This is a simple and effective method to augment the input with more spatial and temporal information, without modifying the architecture~\cite{caesar2020nuscenes}.

However, it has been observed that the effectiveness of fixed aggregation diminishes as more frames are used, eventually degrading the detection performance~\cite{chen2022mppnet,yang20213dman}. Previous works attribute this degradation to the motion of the objects. While the point clouds of stationary objects are inherently aligned after aggregation~\cite{huang2024soap}, producing denser and more complete geometry (\cref{fig:static-dynamic-static}), dynamic object point clouds become misaligned and distorted from motion (\cref{fig:static-dynamic-dynamic}). Yang et al.~\cite{yang20213dman} notice that such misalignment makes multi-frame aggregation unhelpful, even degrading performance for fast-moving objects. Chen et al.~\cite{chen2022mppnet} argue that this effect poses an additional challenge due to different dynamic objects having different distorted point cloud patterns. This introduces a performance trade-off, as illustrated in~\cref{fig:static-dynamic-aph}: the detection of objects at different speeds is optimal using different numbers of frames.

To address this challenge, SOTA multi-frame detectors have made use of attention-based feature-level aggregation to more effectively utilize information from past frames~\cite{chen2022mppnet,yang20213dman,luo2023transpillars}. However, our experiments reveal that there is considerably more performance to be gained from modifying the input, before resorting to specific architectural designs, with the additional complexity and computation costs they entail.

In this work, we therefore propose VADet (Variable Aggregation Detection): a simple and effective alternative to fixed aggregation that sets the level of aggregation of an object according to its properties. As VADet operates at input level, it can be integrated into existing architectures to improve detection performance without significant modifications or computational overhead. Moreover, VADet's low latency supports its use in real-time applications.

The core of VADet is a function $\eta$ that maps each detected object to an empirically-optimal number of frames to aggregate. To construct $\eta$, we propose \emph{random aggregation training} (RAT, \cref{sec:RAT}) to efficiently study the effects of fixed aggregation on detection performance, over a wide range of configurations.

We use RAT to analyze three representative object detection architectures, establishing that in addition to speed, as illustrated in~\cref{fig:static-dynamic}, object point density demonstrates another important trade-off (see~\cref{fig:density_vs_frames}). 
The function $\eta$ is then constructed based on the training data to map an object's estimated speed and point density to a number of frames to aggregate, a process detailed in \cref{sec:va}.

Thanks to the per-object aggregation based on $\eta$, VADet can achieve good performance for objects with different speed and density in the same scene.
Our results (\cref{sec:results}) show that VADet consistently exceeds the performance of fixed aggregation, for a given architecture, and can surpass the performance of much more complex SOTA approaches.

\section{Related Work}
\label{sec:related-work}
Feature-based alignment of sequential point clouds has been explored in 3D object detection. Early work by Luo et al.~\cite{luo2018fastandfurious} uses simple concatenation to combine features from multiple point clouds, which presents a trade-off between accuracy and efficiency depending on the feature layer used for fusion. Naive concatenation of feature maps inevitably introduces misalignment at the feature level due to ego motion. Huang et al.~\cite{huang20203dlstm} instead use an LSTM to encode temporal information as hidden features and address the alignment issue by transforming the feature map using ego motion.

Recently, attention mechanisms have gained popularity in feature fusion and have shown promising results. Yin et al.~\cite{yin2020spatiotemporal} propose a GRU module equipped with spatiotemporal attention for better feature alignment. 3D-MAN~\cite{yang20213dman} and MPPNet~\cite{chen2022mppnet} both employ attention mechanisms to combine features generated from a single or few-frame region proposal network to produce more refined detections. To better utilize the rich multi-scale features, TransPillars~\cite{luo2023transpillars} proposes attention-based feature fusion at the voxel level to preserve the instance and contextual information.

While feature-based multi-frame methods can make effective use of longer temporal input, they often require modifications to the architecture and incur additional computation cost due to the feature transformation and fusion operations. Input-level aggregation, on the other hand, does not require architectural modifications and has been widely adopted by recent work. Caesar et al.~\cite{caesar2020nuscenes} show that directly concatenating multiple consecutive ego-motion-corrected point clouds at the input level can not only improve detection performance but also enable velocity prediction for each detected object, using a velocity regression head. 

We refer to this strategy as fixed aggregation, in contrast to our proposed variable aggregation. Specifically, in fixed aggregation, each point cloud from previous timestamps undergoes ego-motion correction and is then concatenated with the current frame's point cloud.  More formally, let $P_\tau\in\mathbb{R}^{N_\tau\times 3}$ denote $N_\tau$ point coordinates at timestamp $\tau$, with the corresponding ego pose $T_\tau\in\mathbb{SE}(3)$ that represents the transformation from the ego LiDAR coordinate system to a common global coordinate system. Then, the aggregated $n$-frame point cloud at timestamp $\tau$ is defined as
\begin{equation}
    P^*_\tau = \bigoplus_{i=0}^{n-1} P_{\tau-i} (T_\tau^{-1} T_{\tau-i})^\T,
\end{equation}
where $\oplus$ denotes the concatenation operation. Transformation $T_\tau^{-1} T_{\tau-i}$ accounts for the motion of ego between timestamps $\tau$ and $\tau-i$. In addition to spatial coordinates, intensity, and elongation as point features, a separate channel is used to encode relative timestamps.



\section{Method}
\label{sec:method}
VADet addresses the performance trade-off associated with fixed aggregation by adaptively aggregating different types of objects with a different number of frames. To this end, we first introduce Random Aggregation Training (RAT) to enable a single detector to handle a wide range of input frame counts. We then describe our variable aggregation strategy.

\subsection{Random Aggregation Training}\label{sec:RAT}
Studying the impact of the number of input frames on different types of objects (e.g., stationary vs. dynamic) is a crucial component of our approach. Existing works tend to demonstrate the performance trade-offs of aggregation by evaluating multiple detectors trained separately with different fixed frame counts~\cite{yang20213dman,wang2023dsvt,huang2024soap}. This is computationally expensive and thus is often done only for a few configurations. Moreover, we find that assessing performance differences between different input configurations, which can often be subtle, using multiple separately trained models is prone to high variance.

To efficiently explore the effects of frame counts on detection performance, we introduce random aggregation training (RAT), wherein a single detector is trained with input that has randomly varying numbers of aggregated frames per scene. To compensate for the increased variety of the input, we increase the number of training epochs accordingly. We find that even though the model's capacity remains unchanged, RAT allows the model to achieve equivalent or slightly better performance than detectors trained on different fixed configurations. This is demonstrated in~\Cref{tab:rat}. 

\begin{table}[ht]
\centering
\caption{Vehicle AP of a VoxelNeXt~\cite{chen2023voxelnext} detector trained with separate fixed configurations and RAT, evaluated with different input aggregations.}
\label{tab:rat}
\resizebox{0.85\linewidth}{!}{
\begin{tabular}{r|cccc}
\toprule
 & {\bf 3-frame} & {\bf 4-frame} & {\bf 8-frame} & {\bf 16-frame} \\ \midrule
Separate & 72.90 & 73.45 & 74.70 & 75.06 \\
RAT & 73.38 & 74.13 & 75.45 & 75.74 \\ \bottomrule
\end{tabular}
}
\end{table}

RAT thus offers several advantages. In terms of studying the effect of the input aggregation, it significantly reduces the computational cost because training a separate detector for each input configuration is no longer required. This enables us to cover a broader range of frame counts than existing work and to more precisely determine the trade-offs for different types of object. Additionally, as the evaluation is done with a single model and varying input configurations, we find that RAT reduces the variance associated with training, providing us with more consistent results.

In this work, we also use RAT as a pre-training strategy. The detector trained with RAT serves as an ideal starting point for our proposed variable aggregation strategy, thanks to its ability to handle multiple input configurations.

\subsection{Variable Aggregation}
\label{sec:va}
To address the performance trade-off between different types of objects, we propose per-object variable aggregation, which dynamically aggregates each detected object according to its properties, such as speed and point density.

In VADet, we first perform velocity estimation for each detected object. This serves two purposes: first, it allows us to identify the approximate locations of previously detected objects in the current frame using a constant velocity motion model, enabling us to aggregate each region separately using different aggregation strategies; second, it indicates the motion state of the object and is an important factor for determining the number of frames used in the aggregation. To estimate the velocity of the objects, we follow previous methods~\cite{caesar2020nuscenes} and add channels to the regression task representing the $x$ and $y$ components of the velocity vector.

Formally, at timestamp $\tau$, we consider a previously detected bounding box $b_{\tau-1}$
in the coordinate system of the current frame with position $\mathbf{x}_{\tau-1}$, dimensions $(l_{\tau-1},w_{\tau-1},h_{\tau-1})$, heading $\theta_{\tau-1}$, estimated velocity $\mathbf{v}_{\tau-1}$, and $n_{\tau-1}$ points from the point cloud at timestamp $\tau-1$ inside the bounding box. 
To achieve better performance, our strategy is to find a function $\eta(b_{\tau-1})$ that gives the empirically best number of frames to aggregate for each object at the current timestamp $\tau$.

\subsubsection{Learning Function $\eta$}\label{sec:eta}
To determine the number of aggregated frames that provide the best detection performance for each object detection, we consider two important factors: speed and point density (number of points per unit surface area). Aggregation changes the appearance of the point clouds for objects of different speeds due to motion distortion, as illustrated in~\cref{fig:static-dynamic-static} and~\cref{fig:static-dynamic-dynamic}, and thus the optimal number of frames for aggregation varies with the object's speed. Additionally, as more frames are aggregated, the point density increases proportionally, affecting the number of points representing each object and consequently impacting the detection performance.

In practice, since neither of these factors can be accurately determined for a given object, we use the velocity prediction from the object detector to estimate its speed $\lVert v_{\tau-1}\rVert$, and approximate its point density $\rho_{\tau-1}$ using the predicted bounding box dimensions:
\begin{equation}
    \rho_{\tau-1} = n_{\tau-1} / \left(l_{\tau-1}\cdot w_{\tau-1}+l_{\tau-1}\cdot h_{\tau-1}+w_{\tau-1}\cdot h_{\tau-1}\right)
    \label{eq:density}
\end{equation}

Since the trade-offs for different types of objects can be different for each dataset and architecture, we obtain $\eta$ empirically by evaluating the object detector's performance on different types of objects over a wide range of input configurations on the training split. This is feasible thanks to RAT allowing the use of a single model for evaluation. Specifically, we implement $\eta$ as a piecewise function using a lookup table. The table is constructed by dividing the training set into subcategories of objects with different speeds and densities, and determining the frame count that leads to the highest average precision for each subcategory.

While existing works have observed the effects of aggregation on objects with different speeds~\cite{yang20213dman,chen2022mppnet,huang2024soap}, the way aggregation interacts with objects with different point densities is not well-studied. In VADet, we establish point density as an additional factor that should be considered and evaluated.

\subsubsection{Input construction}
For each object, we first determine the approximate location $\hat{\mathbf{x}}^*_{\tau}$ of the object at the current timestamp $\tau$ according to the constant velocity model. This is given by
\begin{equation}
    \hat{\mathbf{x}}^*_{\tau} = \mathbf{x}_{\tau-1} + \mathbf{v}_{\tau-1} / f\label{eq:crop0},
\end{equation}
where $f$ is the frame rate of the LiDAR point clouds.

To encompass all the points belonging to the object, including past points that could potentially fall outside of the bounding box, we enlarge the region of aggregation based on the speed of the object and the number of frames used in the aggregation. The final region of aggregation for an object, denoted $\hat{b}_{\tau}:=(\hat{\mathbf{x}}_{\tau},\hat{l}_{\tau},\hat{w}_{\tau},\hat{h}_{\tau},\hat{\theta}_{\tau})$, is given by
\begin{align}
    \hat{\mathbf{x}}_{\tau} &= \hat{\mathbf{x}}^*_{\tau} - \frac{\mathbf{v}_{\tau-1} \cdot (\eta(b_{\tau-1})-1)}{2f} \label{eq:crop1}, \\
    \hat{l}_{\tau} &= \sigma\cdot l_{\tau-1} + \frac{|\mathbf{v}_{\tau-1}|\cdot(\eta(b_{\tau-1})-1)}{f} \label{eq:crop2}, \\
    \hat{w}_{\tau} &= \sigma\cdot w_{\tau-1} \label{eq:crop3}, \\
    \hat{h}_{\tau} &= \sigma\cdot h_{\tau-1} \label{eq:crop4}, \\
    \hat{\theta}_{\tau} &= \theta_{\tau-1} \label{eq:crop5},
\end{align}
where $\sigma \geq 1$ is the enlargement factor that adds a margin to the aggregation region. Note that the second term in~\cref{eq:crop2} is used to enlarge the length of the object to include the misaligned ``smudges'' from object motion. Accordingly, the center of the aggregation region is adjusted in~\cref{eq:crop1}.

\begin{algorithm}[t]
\caption{Variable Aggregation}
\label{algorithm}
    \vspace{2pt}
    \KwIn{$\begin{cases}
   \text{point clouds } P_{\tau-n_{\max}+1},\dots,P_\tau\\
   \text{ego poses } T_{\tau-n_{\max}+1},\dots,T_\tau\\
   \text{previous detections } \textbf{b}_{\tau-1}
    \end{cases}$}
    \KwOut{aggregated object points $P^{\text{obj}}_{\tau}$\vspace{0.5em}}
    Initialize $P^{\text{obj}}_{\tau}$ to be an empty point cloud\\
    Compute $\hat{\textbf{b}}_{\tau}$ from $\textbf{b}_{\tau-1}$ using~\cref{eq:crop0,eq:crop1,eq:crop2,eq:crop3,eq:crop4,eq:crop5}\\
    \For{$i \gets 0 \mathbf{~to~} n_{\max}-1$}{
      $P^{\text{corr}}_{\tau-i} \gets P_{\tau-i} (T_\tau^{-1} T_{\tau-i})^\T$\\
      $\hat{\textbf{b}}^{i}_{\tau} \gets \left\{\hat{b}_{\tau}\ |\ \hat{b}_{\tau}\in \hat{\textbf{b}}_{\tau}, \eta(b_{\tau-1})>i \right\}$\\
      $P^{\text{obj}}_{\tau} \gets P^{\text{obj}}_{\tau} \oplus \text{Crop}(P^{\text{corr}}_{\tau-i}, \hat{\textbf{b}}^{i}_{\tau})$\\
    }
    \textbf{return} $P^{\text{obj}}_{\tau}$\\[2.5pt]
\end{algorithm}

Finally, for each region constructed from the process above, we aggregate the respective number of past frames in that region. In practice, this operation can be efficiently implemented by cropping the regions $\hat{b}_{\tau}$ for each previous frame $\tau-i$,  $i<\eta(b_{\tau-1})$. Point clouds are corrected for ego motion by transformation $T_\tau^{-1} T_{\tau-i}$ before aggregation. This process is detailed in \cref{algorithm}. The aggregated objects are then combined with the remaining background points outside of the aggregation regions.

\section{Experimental Setup}
\label{sec:setup}

\begin{table*}[t]
\centering
\setlength{\tabcolsep}{0.4em}
\caption{Dynamic vehicle AP of a CenterPoint model on the Waymo dataset using different subset evaluation metrics.}
\resizebox{0.82\textwidth}{!}{
\begin{tabular}{r|c|*{11}{>{\centering}p{0.8cm}}c}
\toprule
\multirow{2}{*}{\bf Metric} & {\bf Dynamic} & \multicolumn{11}{c}{{\bf Subsets} (m/s)} & {\bf Weighted} \\
 & $\geq$0.2\,m/s & [0.2,1) & [1,3) & [3,5) & [5,7) & [7,9) & [9,11) & [11,13) & [13,15) & [15,17) & [17,20) & $\geq$20 & {\bf average} \\ \midrule
Waymo & 72.9 & 63.0 & 62.4 & 62.9 & 66.5 & 68.9 & 66.8 & 64.1 & 61.8 & 64.0 & 67.0 & 62.9 & 64.7 \\
Ours & 74.4 & 71.6 & 69.8 & 72.2 & 74.6 & 76.6 & 76.3 & 75.7 & 75.3 & 78.3 & 81.6 & 78.3 & 74.4 \\ \bottomrule
\end{tabular}
}
\label{tab:subset_ap}
\end{table*}

To demonstrate our method is effective and can be easily applied to different architectures, we evaluate VADet using CenterPoint~\cite{yin2021center}, VoxelNeXt~\cite{chen2023voxelnext}, and DSVT~\cite{wang2023dsvt} on the large scale Waymo Open Dataset~\cite{sun2020waymo}. CenterPoint, VoxelNeXt, and DSVT respectively represent the SOTA in dense voxel-based, fully sparse, and transformer-based 3D object detectors.

\subsection{Dataset}
The Waymo dataset is a large-scale autonomous driving dataset collected under a variety of traffic conditions in San Francisco, Phoenix, and Mountain View. It consists of 798 sequences for the training split, 202 sequences for the validation split, and 150 sequences for the held-back test split. Each sequence is approximately 20 seconds long.

Waymo uses multiple LiDAR sensors operating at 10\,Hz, resulting in approximately 200 point clouds per sequence. The main sensor is a top-mounted proprietary 64-beam rotating LiDAR. There are also four close-range LiDARs mounted to the side of the vehicle. In addition to the intensity channel, Waymo's sensors also produce elongation for each point. We use points from all five LiDARs by concatenating them in the ego vehicle coordinate system.

\subsection{Evaluation Metrics}
For overall object detection performance, we use the official Waymo evaluation suite. Unless specified otherwise, we report the level 2 average precision (AP) for Vehicle, which includes very sparse objects ($\leq5$ points). The IoU threshold used for matching true positives is 0.7.

Evaluating a specific subset of the objects (e.g., dynamic objects) for more detailed analysis requires more careful handling due to false positives that cannot be matched with any ground truth. Waymo evaluation suite provides functionalities to perform such evaluation. Specifically, the subset precision and recall are defined as follows:
\begin{align}
    \mathrm{Prec}_{\mathrm{waymo}} &= \frac{\mathrm{TP}_{\mathrm{subset}}}{\mathrm{TP}_{\mathrm{subset}} + \mathrm{FP}_{\mathrm{subset}} + \mathrm{FP}_{\mathrm{unknown}}}, \\
    \mathrm{Rec}_{\mathrm{waymo}} &= \frac{\mathrm{TP}_{\mathrm{subset}}}{\mathrm{TP}_{\mathrm{subset}} + \mathrm{FN}_{\mathrm{subset}}}.
\end{align}
$\mathrm{TP}_{\mathrm{subset}}$ and $\mathrm{FN}_{\mathrm{subset}}$ are the number of true positives and false negatives within the subset of objects. $\mathrm{FP}_{\mathrm{subset}}$ is the number of false positives that overlap with some objects in the subset but do not meet the IoU threshold. $\mathrm{FP}_{\mathrm{unknown}}$ is the number of objects that do not overlap with any ground truth.

However, we argue that this formulation cannot correctly reflect the detection performance of a subset because $\mathrm{FP}_{\mathrm{unknown}}$ is independent of the subset and biases the precision depending on the size of the subset.
The consequence is that subsets of different sizes are incomparable and the weighted sum of subset precisions underestimates the precision of the union of subsets.

To address this issue, we introduce a definition of precision that weights $\mathrm{FP}_{\mathrm{unknown}}$ by the proportion of objects in the subset:
\begin{equation}
    \mathrm{Prec}_{\mathrm{subset}} = \frac{\mathrm{TP}_{\mathrm{subset}}}{\mathrm{TP}_{\mathrm{subset}} + \mathrm{FP}_{\mathrm{subset}} + \frac{N_{\mathrm{subset}}}{N_{\mathrm{total}}}\cdot\mathrm{FP}_{\mathrm{unknown}}}
\end{equation}
$N_{\mathrm{subset}}$ and $N_{\mathrm{total}}$ are the number of objects in the subset and the total number of objects, respectively. The definition of recall remains unchanged. In contrast to the standard Waymo metric, this formulation allows the comparison and combination of subset APs, as illustrated in~\Cref{tab:subset_ap}. We note, in particular, that when using our formulation the weighted average of subset APs weighted by subset size is closer to the AP of the union of the subsets.

\subsection{Implementation Details}
\subsubsection{Architectures}

CenterPoint~\cite{yin2021center} and VoxelNeXt~\cite{chen2023voxelnext} are both single stage CNN-based 3D object detectors. The input point cloud first undergoes voxelization and is then fed to a sparse convolutional backbone. Multiple detection heads are used to separately produce bounding box attributes, including confidence score, location, and box dimensions.
DSVT~\cite{wang2023dsvt} is an emerging transformer-based 3D object detector. In DSVT, the traditional convolution backbone is replaced with multiple transformer blocks consisting of shifted window and partition-based self-attention operations.

Following~\cite{caesar2020nuscenes}, we add a two-layer regression head to predict an object's velocity vector for all architectures. For DSVT, we use the pillar variant (which we denote DSVT-P). As the dynamic voxelization adopted by the original work cannot be scaled beyond 4-frame aggregation on our hardware, we use a traditional static voxelization where for each voxel, at most 40 points are randomly selected and processed. To make the computation tractable, we further reduce the input channels from 192 to 96, and the hidden channels from 384 to 192.

\begin{table*}[t]
\centering
\setlength{\tabcolsep}{0.4em}
\caption{Overall vehicle AP on the Waymo validation split.
The best performance is in bold and the second best is underlined.}
\label{tab:results-overall}
\resizebox{0.75\textwidth}{!}{%
\begin{tabular}{c|c|llllllllllllll}
\toprule
 & \textbf{VADet} & \multicolumn{1}{c}{\textbf{3f}} & \multicolumn{1}{c}{\textbf{4f}} & \multicolumn{1}{c}{\textbf{5f}} & \multicolumn{1}{c}{\textbf{6f}} & \multicolumn{1}{c}{\textbf{7f}} & \multicolumn{1}{c}{\textbf{8f}} & \multicolumn{1}{c}{\textbf{9f}} & \multicolumn{1}{c}{\textbf{10f}} & \multicolumn{1}{c}{\textbf{11f}} & \multicolumn{1}{c}{\textbf{12f}} & \multicolumn{1}{c}{\textbf{13f}} & \multicolumn{1}{c}{\textbf{14f}} & \multicolumn{1}{c}{\textbf{15f}} & \multicolumn{1}{c}{\textbf{16f}} \\ \midrule
CenterPoint & \textbf{71.5} & 68.1 & 68.9 & 69.5 & 70.0 & 70.3 & 70.5 & 70.7 & 70.8 & 70.9 & {\ul 71.0} & {\ul 71.0} & {\ul 71.0} & {\ul 71.0} & 70.9 \\
VoxelNeXt & \textbf{76.5} & 73.4 & 74.1 & 74.6 & 75.0 & 75.3 & 75.4 & 75.6 & 75.7 & 75.7 & {\ul 75.8} & {\ul 75.8} & {\ul 75.8} & {\ul 75.8} & 75.7 \\
DSVT-P & \textbf{74.5} & 69.5 & 70.3 & 70.8 & 71.4 & 71.7 & 72.0 & 72.2 & 72.4 & 72.6 & 72.7 & 72.8 & 73.0 & 73.0 & {\ul 73.2} \\ \bottomrule
\end{tabular}%
}
\end{table*}

\subsubsection{Training}
The baseline models are trained on the entire Waymo training split for 20 epochs across 8 NVIDIA A6000 GPUs using the proposed RAT strategy. We use the Adam optimizer and a one-cycle learning rate schedule, with an initial learning rate of 0.0003 and a maximum learning rate of 0.003. We use a total batch size of 32 for CenterPoint, 32 for VoxelNeXt, and 16 for DSVT-P.

We initialize our VADet models with the weights from the baseline models, then fine-tune them for an additional epoch using cosine learning rate decay with an initial learning rate of 0.0001. Since our models rely on information cached from previous frames, including point clouds and predictions, the standard frame-based shuffling cannot be applied during training. Instead, we divide the dataset into mini-sequences with a maximum of 32 frames and shuffle the mini-sequences. This introduces randomness while ensuring that frames appear in their correct sequence. Furthermore, as the performance of a model can fluctuate during training (due to ongoing optimization), following~\cite{fan2023fsd++}, we load the offline predictions from the baseline models using 3-frame fixed aggregation during training.

\subsubsection{Specifying $\eta$}
We implement $\eta$ as a lookup table: we subdivide the training dataset based on speed and point cloud density and empirically determine the frame count that leads to the highest average precision for each subcategory (\cref{sec:va}). The speed and density thresholds are set to [0.00, 0.20, 1.55, 3.63, 5.90, 81.6, 11.34, 17.53]\,m/s and [0.00, 0.68, 1.86, 3.86, 8.02, 18.81, 71.37] pts/m$^2$ respectively. They are chosen based on the training set object statistics to ensure a sufficient number of objects in each bin for evaluation. For each speed and density combination, we evaluate the baseline model on 3--16-frame input and select the frame count with the highest AP performance. Background points undergo a fixed 3-frame aggregation.

\begin{table}[t]
\centering
\caption{Overall vehicle performance on the Waymo validation split compared with SOTA multi-frame detectors.}
\resizebox{0.9\linewidth}{!}{%
\begin{tabular}{l|c|c|c}
\toprule
\textbf{Method} & \textbf{\# frames} & \textbf{L1 AP/APH} & \textbf{L2 AP/APH} \\ \midrule
PillarNeXt-B~\cite{li2023pillarnext} & 3 & 80.6/80.1 & 72.9/72.4 \\
DSVT-pillar~\cite{wang2023dsvt} & 4 & 81.7/81.2 & 73.8/73.4 \\
DSVT-voxel~\cite{wang2023dsvt} & 4 & 81.8/81.4 & 74.1/73.6 \\
FSD++~\cite{fan2023fsd++} & 7 & 81.4/80.9 & 73.3/72.9 \\
CenterFormer~\cite{zhou2022centerformer} & 8 & 78.8/78.3 & 74.3/73.8 \\
3D-MAN~\cite{yang20213dman} & 16 & 74.5/74.0 & 67.6/67.1 \\
MPPNet~\cite{chen2022mppnet} & 16 & {\ul 82.7/82.3} & {\ul 75.4/75.0} \\ \midrule
VADet-CenterPoint & 3--16 & 79.0/78.5 & 71.7/71.3 \\
VADet-DSVT-P & 3--16 & 82.0/81.6 & 74.5/74.1 \\ 
VADet-VoxelNeXt & 3--16 & \textbf{83.9/83.4} & \textbf{76.6/76.1} \\ \bottomrule
\end{tabular}%
}
\label{tab:results-sota}
\end{table}

\begin{table}[t]
\centering
\caption{Overall vehicle performance on the Waymo test split compared with other methods (without TTA or ensemble).}
\resizebox{0.9\linewidth}{!}{%
\begin{tabular}{l|c|c|c}
\toprule
\textbf{Method} & \textbf{\# frames} & \textbf{Modality} & \textbf{L2 AP/APH} \\ \midrule
AFDetV2~\cite{hu2022afdetv2} & 2 & L & 74.3/73.9 \\
PV-RCNN++~\cite{shi2023pvrcnn++} & 2 & L & 76.3/75.9 \\
SWFormer~\cite{sun2022swformer} & 3 & L & 75.0/74.7 \\
PillarNeXt-B~\cite{li2023pillarnext} & 3 & L & 76.2/75.8 \\
FSD++~\cite{fan2023fsd++} & 7 & L & 77.1/76.7 \\
3D-MAN~\cite{yang20213dman} & 16 & L & 70.4/70.0 \\
MPPNet~\cite{chen2022mppnet} & 16 & L & 77.3/76.9 \\
CenterFormer~\cite{zhou2022centerformer} & 16 & L & {\ul 78.7/78.3} \\ \midrule
BEVFusion~\cite{liu2023bevfusion} & 3 & C+L & 77.9/77.5 \\
DeepFusion~\cite{li2022deepfusion} & 5 & C+L & 76.1/75.7 \\
HorizonLiDAR3D~\cite{ding20201st} & 5 & C+L & 78.2/77.8  \\
LoGoNet~\cite{li2023logonet} & 5 & C+L & {\ul 79.7/79.3} \\ \midrule
VADet-VoxelNeXt & 3--16 & L & \textbf{79.8/79.4} \\ \bottomrule
\end{tabular}%
}
\label{tab:results-sota-test}
\end{table}


\section{Results and Analysis}
\label{sec:results}

We present the overall performance of VADet compared to baseline models and SOTA methods in~\Cref{sec:results-overall}. To demonstrate the better trade-offs achieved by VADet, in~\Cref{sec:results-breakdown} we perform a detailed evaluation based on speed and point cloud density. 

\subsection{Overall Performance}
\label{sec:results-overall}

\Cref{tab:results-overall} shows that, for all three architectures, VADet demonstrates superior performance compared to baseline models using different fixed frame counts. In particular, VADet using CenterPoint achieves 71.5 AP compared to 71.0 AP using 12, 13, 14, or 15-frame aggregation. For VoxelNeXt, VADet achieves 76.5 AP compared to 75.8 AP using 12, 13, 14, or 15-frame aggregation. DSVT-P with VADet achieves 74.5 AP compared to 73.2 AP using 16-frame aggregation.

When compared with SOTA object detection methods in~\Cref{tab:results-sota} and~\Cref{tab:results-sota-test}, VADet-equipped models also have competitive performance on both the validation and test splits. Most notably, our single-stage VADet-VoxelNeXt achieves 76.1 and 79.4 level 2 APH on the validation and test split respectively, outperforming two-stage multi-frame methods such as MPPNet~\cite{chen2022mppnet} by a large margin, with lower computation overhead. Specifically, we measure that the second stage proposed by MPPNet introduces an additional 900--2500\,ms latency over the base detector, while VADet only requires an additional 50\,ms overhead for input aggregation, which we believe can be further optimized with a GPU-accelerated implementation. Furthermore, with a powerful backbone such as VoxelNeXt, our method can match the performance of recent SOTA camera-LiDAR fusion methods such as LoGoNet~\cite{li2023logonet}.

\begin{figure}[t]
    \centering
    \begin{subfigure}[b]{0.32\linewidth}
         \centering
         \includegraphics[width=\linewidth]{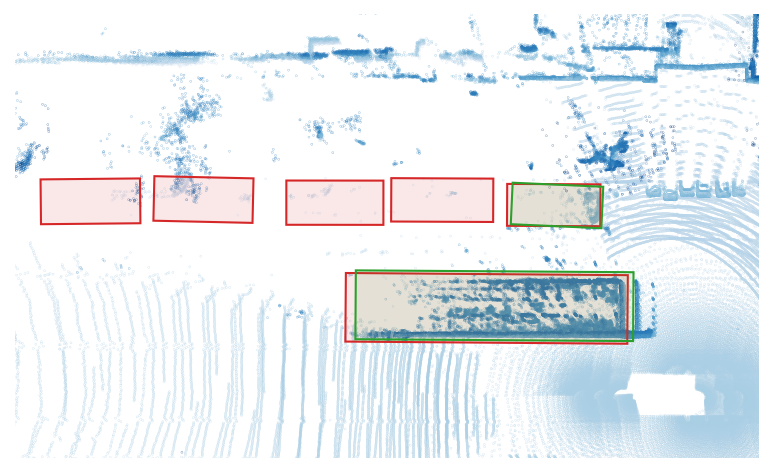}
     \end{subfigure}
     \begin{subfigure}[b]{0.32\linewidth}
         \centering
         \includegraphics[width=\linewidth]{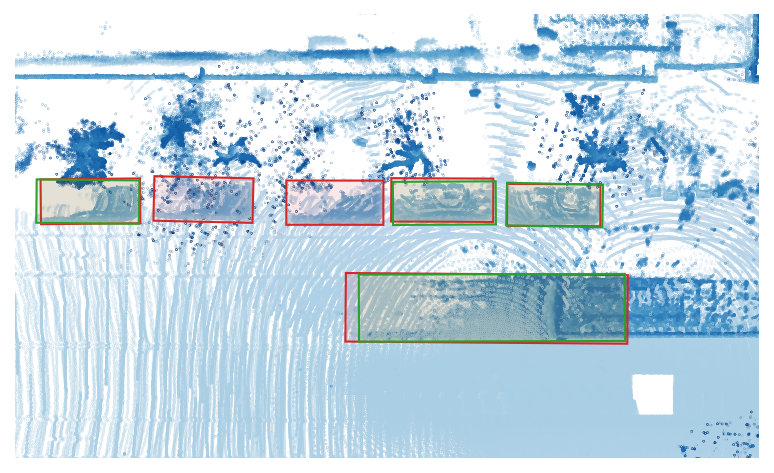}
     \end{subfigure}
     \begin{subfigure}[b]{0.32\linewidth}
         \centering
         \includegraphics[width=\linewidth]{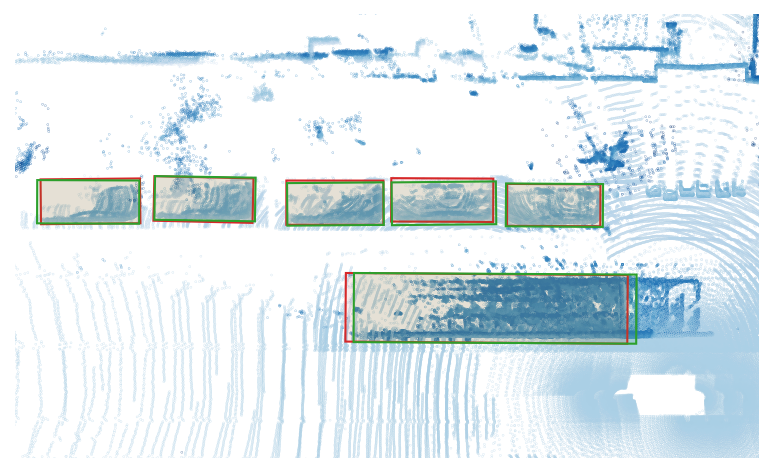}
     \end{subfigure}
     \begin{subfigure}[b]{0.32\linewidth}
         \centering
         \includegraphics[width=\linewidth]{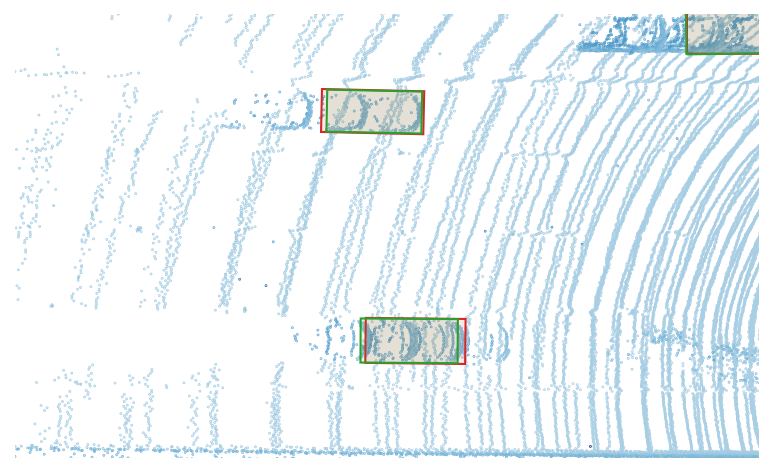}
         \caption{3-frame}
     \end{subfigure}
     \begin{subfigure}[b]{0.32\linewidth}
         \centering
         \includegraphics[width=\linewidth]{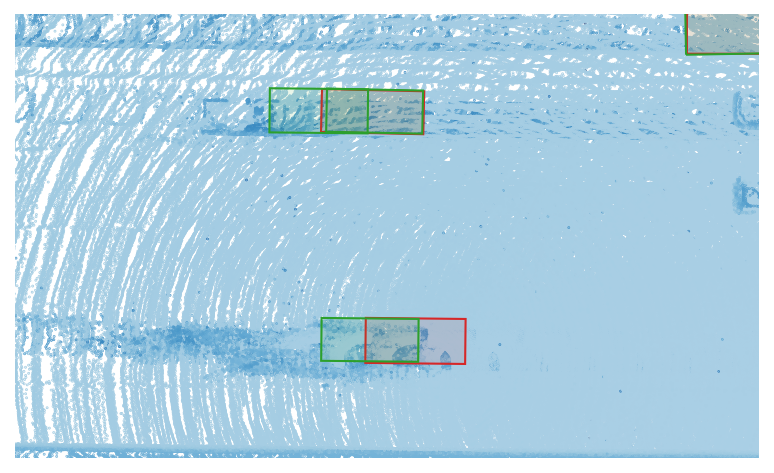}
         \caption{16-frame}
     \end{subfigure}
     \begin{subfigure}[b]{0.32\linewidth}
         \centering
         \includegraphics[width=\linewidth]{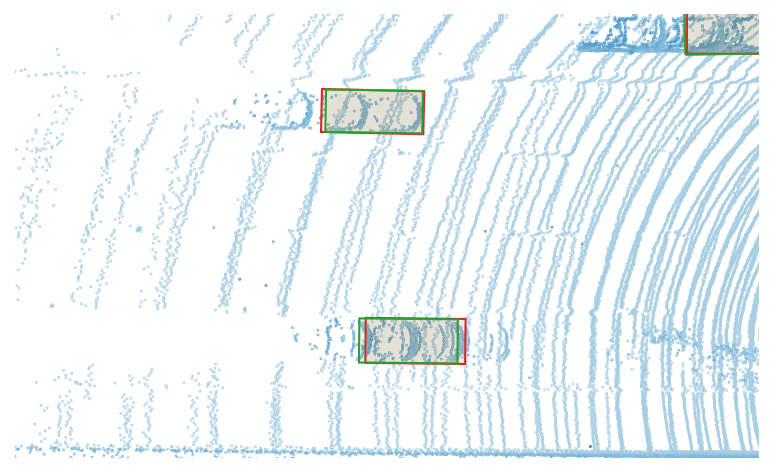}
         \caption{VADet}
     \end{subfigure}
    \caption{Qualitative results comparing 3-frame and 16-frame fixed aggregation with VADet. Red and green bounding boxes are ground truth and predictions, respectively. Predictions are filtered with 0.5 confidence threshold for visual clarity.}
    \label{fig:qualitative}
\end{figure}

\begin{table*}[t]
\centering
\setlength{\tabcolsep}{0.4em}
\caption{Vehicle AP breakdown by speed.} 
\label{tab:results-speed}
\resizebox{0.82\textwidth}{!}{%
\begin{tabular}{c|c|c|llllllllllllll}
\toprule
 & \textbf{Speed} & \textbf{VADet} & \multicolumn{1}{c}{\textbf{3f}} & \multicolumn{1}{c}{\textbf{4f}} & \multicolumn{1}{c}{\textbf{5f}} & \multicolumn{1}{c}{\textbf{6f}} & \multicolumn{1}{c}{\textbf{7f}} & \multicolumn{1}{c}{\textbf{8f}} & \multicolumn{1}{c}{\textbf{9f}} & \multicolumn{1}{c}{\textbf{10f}} & \multicolumn{1}{c}{\textbf{11f}} & \multicolumn{1}{c}{\textbf{12f}} & \multicolumn{1}{c}{\textbf{13f}} & \multicolumn{1}{c}{\textbf{14f}} & \multicolumn{1}{c}{\textbf{15f}} & \multicolumn{1}{c}{\textbf{16f}} \\ \midrule
\multirow{3}{*}{CenterPoint} & Stationary & \textbf{70.3} & \multicolumn{1}{c}{65.8} & \multicolumn{1}{c}{66.8} & \multicolumn{1}{c}{67.6} & \multicolumn{1}{c}{68.1} & \multicolumn{1}{c}{68.5} & \multicolumn{1}{c}{68.8} & \multicolumn{1}{c}{69.1} & \multicolumn{1}{c}{69.3} & \multicolumn{1}{c}{69.4} & \multicolumn{1}{c}{69.5} & \multicolumn{1}{c}{69.6} & \multicolumn{1}{c}{69.6} & \multicolumn{1}{c}{{\ul 69.7}} & \multicolumn{1}{c}{{\ul 69.7}} \\
 & Slow & \textbf{75.0} & \multicolumn{1}{c}{74.1} & \multicolumn{1}{c}{74.5} & \multicolumn{1}{c}{74.7} & \multicolumn{1}{c}{{\ul 74.9}} & \multicolumn{1}{c}{{\ul 74.9}} & \multicolumn{1}{c}{74.8} & \multicolumn{1}{c}{74.8} & \multicolumn{1}{c}{74.7} & \multicolumn{1}{c}{74.6} & \multicolumn{1}{c}{74.6} & \multicolumn{1}{c}{74.5} & \multicolumn{1}{c}{74.4} & \multicolumn{1}{c}{74.2} & \multicolumn{1}{c}{74.0} \\
 & Fast & \textbf{79.5} & \multicolumn{1}{c}{78.9} & \multicolumn{1}{c}{79.0} & \multicolumn{1}{c}{{\ul 79.1}} & \multicolumn{1}{c}{79.0} & \multicolumn{1}{c}{78.9} & \multicolumn{1}{c}{78.9} & \multicolumn{1}{c}{78.8} & \multicolumn{1}{c}{78.6} & \multicolumn{1}{c}{78.3} & \multicolumn{1}{c}{77.9} & \multicolumn{1}{c}{77.8} & \multicolumn{1}{c}{77.4} & \multicolumn{1}{c}{77.0} & \multicolumn{1}{c}{76.6} \\ \midrule
\multirow{3}{*}{VoxelNeXt} & Stationary & \textbf{74.3} & 70.4 & 71.3 & 71.9 & 72.3 & 72.6 & 72.9 & 73.1 & 73.2 & 73.3 & 73.4 & 73.5 & 73.5 & {\ul 73.6} & {\ul 73.6} \\
 & Slow & \textbf{79.6} & 78.5 & 78.9 & 79.2 & 79.3 & {\ul 79.4} & {\ul 79.4} & 79.3 & 79.3 & 79.2 & 79.2 & 79.1 & 78.9 & 78.8 & 78.7 \\
 & Fast & \textbf{84.2} & 83.6 & 83.8 & 83.8 & {\ul 83.9} & 83.8 & 83.7 & 83.6 & 83.5 & 83.3 & 83.2 & 83.0 & 82.6 & 82.4 & 81.9 \\ \midrule
\multirow{3}{*}{DSVT-P} & Stationary & \textbf{72.3} & 66.4 & 67.4 & 68.1 & 68.7 & 69.1 & 69.5 & 69.8 & 70.1 & 70.3 & 70.6 & 70.7 & 70.9 & 71.0 & {\ul 71.2} \\
 & Slow & \textbf{77.3} & 75.0 & 75.4 & 75.5 & 75.6 & {\ul 75.7} & 75.6 & 75.5 & 75.5 & 75.5 & 75.5 & 75.5 & 75.4 & 75.3 & 75.3 \\
 & Fast & \textbf{82.6} & {\ul 81.2} & {\ul 81.2} & 81.0 & 80.9 & 80.8 & 80.7 & 80.7 & 80.6 & 80.6 & 80.5 & 80.5 & 80.4 & 80.3 & 80.2 \\ \bottomrule
\end{tabular}%
}
\end{table*}

\begin{figure*}[t]
    \centering
    \begin{subfigure}[b]{0.3\textwidth}
         \centering
         \includegraphics[width=\textwidth]{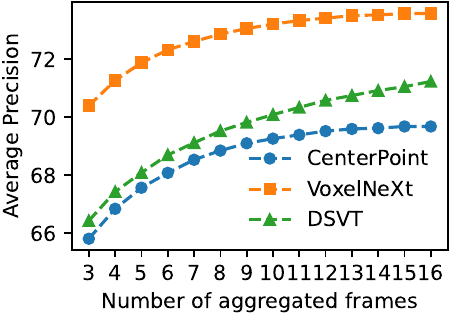}
         \caption{Stationary}
         \label{fig:stationary_vs_frames}
     \end{subfigure}
     \begin{subfigure}[b]{0.3\textwidth}
         \centering
         \includegraphics[width=\textwidth]{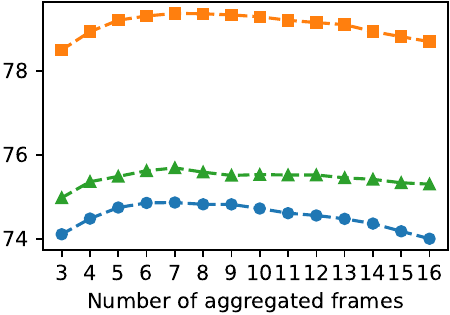}
         \caption{Slow}
         \label{fig:slow_vs_frames}
     \end{subfigure}
     \begin{subfigure}[b]{0.3\textwidth}
         \centering
         \includegraphics[width=\textwidth]{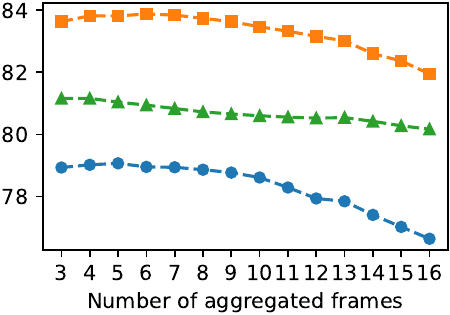}
         \caption{Fast}
         \label{fig:fast_vs_frames}
     \end{subfigure}
    \caption{AP vs. the number of frames for stationary ($<$0.2\,m/s), slow ([0.2,10)\,m/s), and fast-moving ($\geq$10\,m/s) vehicles.}
    \label{fig:speed_vs_frames}
\end{figure*}

Our results demonstrate that VADet can effectively utilize multiple frames to achieve SOTA detection performance, suggesting that by addressing various performance trade-offs with carefully constructed input, a simple single-stage object detection architecture such as VoxelNeXt can outperform much more complex SOTA methods.


\begin{table*}[t]
\centering
\setlength{\tabcolsep}{0.4em}
\caption{Dynamic vehicle AP performance breakdown by point cloud density.}
\label{tab:results-density}
\resizebox{0.82\textwidth}{!}{%
\begin{tabular}{c|c|c|cccccccccccccc}
\toprule
 & \textbf{Density} & \textbf{VADet} & \multicolumn{1}{c}{\textbf{3f}} & \multicolumn{1}{c}{\textbf{4f}} & \multicolumn{1}{c}{\textbf{5f}} & \multicolumn{1}{c}{\textbf{6f}} & \multicolumn{1}{c}{\textbf{7f}} & \multicolumn{1}{c}{\textbf{8f}} & \multicolumn{1}{c}{\textbf{9f}} & \multicolumn{1}{c}{\textbf{10f}} & \multicolumn{1}{c}{\textbf{11f}} & \multicolumn{1}{c}{\textbf{12f}} & \multicolumn{1}{c}{\textbf{13f}} & \multicolumn{1}{c}{\textbf{14f}} & \multicolumn{1}{c}{\textbf{15f}} & \multicolumn{1}{c}{\textbf{16f}} \\ \midrule
\multirow{3}{*}{CenterPoint} & Sparse & \textbf{27.6} & 24.4 & 25.6 & 26.3 & 26.5 & {\ul 26.6} & 26.4 & 26.4 & 26.1 & 25.8 & 25.6 & 25.2 & 24.9 & 24.4 & 23.9 \\
 & Medium & \textbf{87.8} & 87.5 & 87.5 & {\ul 87.6} & {\ul 87.6} & {\ul 87.6} & {\ul 87.6} & 87.5 & 87.4 & 87.3 & 87.1 & 87.1 & 86.9 & 86.7 & 86.5 \\
 & Dense & \textbf{98.9} & \textbf{98.9} & \textbf{98.9} & \textbf{98.9} & \textbf{98.9} & {\ul 98.8} & {\ul 98.8} & {\ul 98.8} & {\ul 98.8} & 98.7 & 98.7 & 98.7 & 98.6 & 98.5 & 98.5 \\ \midrule
\multirow{3}{*}{VoxelNeXt} & Sparse & \textbf{32.8} & 29.8 & 30.8 & 31.6 & 32.0 & {\ul 32.2} & 32.0 & 31.9 & 31.6 & 31.2 & 31.0 & 30.8 & 30.1 & 29.8 & 29.3 \\
 & Medium & \textbf{91.5} & 91.0 & 91.2 & {\ul 91.3} & 91.2 & {\ul 91.3} & 91.2 & 91.2 & 91.1 & 91.1 & 91.0 & 91.0 & 90.8 & 90.7 & 90.5 \\
 & Dense & {\ul 99.3} & \textbf{99.4} & \textbf{99.4} & \textbf{99.4} & \textbf{99.4} & \textbf{99.4} & 99.3 & 99.3 & 99.3 & 99.3 & 99.2 & 99.2 & 99.2 & 99.3 & 99.2 \\ \midrule
\multirow{3}{*}{DSVT-P} & Sparse & \textbf{28.0} & 24.8 & 25.6 & 26.0 & 26.2 & 26.3 & 26.3 & {\ul 26.4} & {\ul 26.4} & 26.3 & {\ul 26.4} & 26.3 & 26.2 & 26.1 & 26.0 \\
 & Medium & \textbf{89.9} & 88.1 & {\ul 88.2} & {\ul 88.2} & {\ul 88.2} & 88.1 & 88.0 & 87.9 & 87.9 & 87.8 & 87.8 & 87.8 & 87.7 & 87.6 & 87.5 \\
 & Dense & \textbf{99.0} & {\ul 98.8} & 98.7 & 98.7 & 98.7 & 98.6 & 98.5 & 98.5 & 98.4 & 98.4 & 98.3 & 98.3 & 98.2 & 98.0 & 98.1 \\ \bottomrule
\end{tabular}%
}
\end{table*}

\begin{figure*}[t]
    \centering
    \begin{subfigure}[b]{0.3\textwidth}
         \centering
         \includegraphics[width=\textwidth]{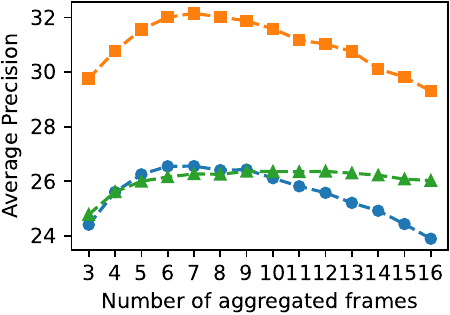}
         \caption{Sparse}
         \label{fig:sparse_vs_frames}
     \end{subfigure}
     \begin{subfigure}[b]{0.3\textwidth}
         \centering
         \includegraphics[width=\textwidth]{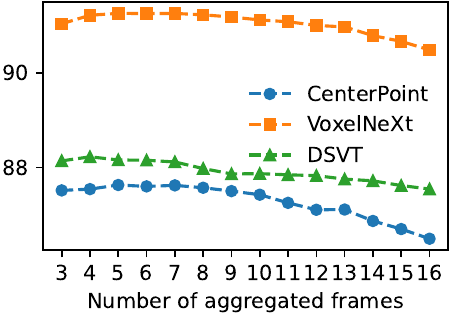}
         \caption{Medium}
         \label{fig:medium_vs_frames}
     \end{subfigure}
     \begin{subfigure}[b]{0.3\textwidth}
         \centering
         \includegraphics[width=\textwidth]{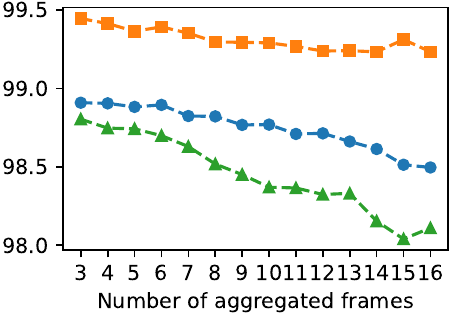}
         \caption{Dense}
         \label{fig:dense_vs_frames}
     \end{subfigure}
    \caption{AP vs. the number of frames for sparse ($<$2\,pts/m$^2$), medium ((2,100]\,pts/m$^2$), and dense ($>$100 pts/m$^2$) dynamic vehicles.}
    \label{fig:density_vs_frames}
\end{figure*}

\subsection{Qualitative Results}
\label{sec:results-qualitative}

Qualitative results comparing 3-frame and 16-frame fixed aggregation with VADet are shown in~\cref{fig:qualitative}. Overall, we observe that 16-frame input appears much denser than 3-frame and VADet. This is especially noticeable for background points, which may not be beneficial to detecting objects and introduce unnecessary computation.

The 3-frame model is unable to detect the occluded stationary vehicles depicted in the upper images, due to the lack of points. While these objects appear more complete with 16 frames, the model can only partially detect them. We hypothesize this is due to the dense background points leading to confusion. VADet, on the other hand, aggregates only object points while keeping the background sparse, and therefore can accurately detect all vehicles.

Similarly, for the fast-moving vehicles seen in the lower images, 16-frame aggregation results in long ``smudges'' around each vehicle, leading to inaccurate localization. In these situations, VADet does not over-aggregate these objects and benefits from the more accurate detections achieved by 3-frame input.

\subsection{Breakdown Analysis}
\label{sec:results-breakdown}

The overall performance degradation at 16-frame fixed aggregation, if any, appears to be minuscule for the baseline models: just 0.1\% degradation for CenterPoint and VoxelNeXt. This is due to the extreme imbalance between different types of objects in the dataset. For instance, the overall results are skewed in favour of the approximately 80\% stationary vehicles in the dataset, which negatively affects the performance of dynamic objects. As a result, we perform a breakdown analysis to highlight the performance trade-offs and demonstrate how VADet can benefit objects that are negatively impacted by fixed aggregation.

\subsubsection{Speed}
To illustrate the impact of input aggregation on objects with different speeds, we divide the vehicle class into stationary ($<$0.2\,m/s), slow ([0.2,10)\,m/s), and fast ($\geq$10\,m/s) subcategories. Stationary, slow, and fast vehicles respectively amount to 79.7\%, 14.2\%, and 6.1\% of the Waymo validation set. The AP performance for each subcategory is reported in~\Cref{tab:results-speed}.

Our results for the baseline models are consistent with the trade-off observed in existing work and further suggest that this effect is consistent across different architectures. Moreover, the results show that the optimal number of frames can be different for objects with different speeds. Specifically, the stationary vehicle performance (\cref{fig:stationary_vs_frames}) increases as more frames are used for aggregation: all baseline models achieve the best performance with 16-frame input, suggesting more aggregation is beneficial. Slow vehicles (\cref{fig:slow_vs_frames}), on the other hand, reach maximum performance at 7 frames, while for fast-moving vehicles (\cref{fig:fast_vs_frames}), a large degradation can be observed when more frames are used, indicating aggregating fewer frames is more favourable.

VADet, on the other hand, does not exhibit such a performance trade-off. For all three architectures and subcategories, VADet achieves higher AP than the best fixed aggregation configuration in each respective subcategory (underlined in~\Cref{tab:results-speed}). This not only highlights the effectiveness of VADet at mitigating the performance trade-off between objects at different speeds but also demonstrates the applicability of VADet to various architectures.

\subsubsection{Point density}
To demonstrate that input aggregation can also lead to a performance trade-off between objects with different point cloud densities, we evaluate and report the performance of vehicles with different point cloud densities. Using~\cref{eq:density}, we divide the vehicle class into sparse ($<$2\,pts/m$^2$), medium ([2,100)\,pts/m$^2$), and dense ($\geq$100\,pts/m$^2$) subcategories.

For stationary objects, we observe the same trend seen in~\cref{fig:stationary_vs_frames}, suggesting the performance is not influenced by the point cloud density. The following analysis therefore focuses on dynamic objects ($\geq$0.2\,m/s), with the full results for stationary objects in the supplementary material. The dynamic vehicles in the validation set contain 23.9\% sparse, 67.9\% medium, and 8.2\% dense vehicles according to our definition. The AP performance for each subcategory is detailed in~\Cref{tab:results-density}.

While we have previously observed that dynamic objects favour fewer input frame counts, we notice in~\cref{fig:sparse_vs_frames} that sparse objects can still benefit from more input aggregation: up to 7-frame aggregation for CenterPoint and VoxelNeXt, and up to 12-frame aggregation for DSVT-P. On the other hand, for denser objects (\cref{fig:medium_vs_frames,fig:dense_vs_frames}), higher frame counts become harmful to the detection performance. All three architectures achieve the best performance for medium density vehicles with 5-frame input, and dense vehicles with 3-frame input. Our results highlight a trade-off that has not been studied in existing work and underscores the necessity of considering point density in our proposed approach.

Compared to fixed aggregation, VADet is comparable to the best performance in each subcategory, surpassing fixed aggregation in many cases. This demonstrates that VADet is also effective at addressing the trade-off between objects with different point densities for different architectures.

\section{Limitations and Extensions}
\label{sec:limitations}
\vspace{-0.2em}

Input aggregation adds information to sparse detections, up to the point that an object's motion or other characteristics cause confusion. VADet therefore improves the detection of certain objects by not over-aggregating them, but their detections may nevertheless be sparse and could benefit from aggregation. For such objects, the addition of a different aggregation approach would be necessary.

In this work, we have identified speed and point density as important features to provide as inputs to function $\eta$. We hypothesize that a future implementation of $\eta$ as a more abstract learned function from point clouds to numbers of frames will produce even better results.

In the preceding text, we have focused on the Waymo vehicle class---a common expedient to simplify comparisons. In the supplementary material, we give further results that show VADet is also effective on the Waymo pedestrian class. While VADet's benefits will be class and dataset dependent, we do not anticipate any obvious limitations.

\section{Conclusion}
\label{sec:conclusion}
\vspace{-0.2em}

We have addressed the inherent performance trade-off of fixed aggregation by proposing VADet, a variable aggregation approach that can be easily applied to different architectures with minimal modifications. Our extensive evaluation shows that VADet can effectively combat the trade-off and achieve SOTA performance. 

\bibliographystyle{ieee_fullname}
\bibliography{main}




\end{document}